\documentstyle[aclap]{article}

\author{Beryl Hoffman \\
Centre for Cognitive Science \\ University of Edinburgh \\
2 Buccleuch Place \\ 
Edinburgh, EH8 9LW, U.K. \\ 
{\tt hoffman@cogsci.ed.ac.uk}
}

\date{}

\title{Translating into Free Word Order Languages}

\newcommand{\gloss}[4]{
\begin{tabular}[t]{@{}*{#1}{l@{\ }}}
#2\\ #3\\ \multicolumn{#1}{@{}l@{}}{#4} \\
\end{tabular}}
\newcommand{\lessgloss}[3]
{
\begin{tabular}[t]{@{}*{#1}{l@{\ }}}
#2\\ #3 \\[2ex]
\end{tabular}}
\newcommand{\extragloss}[5]
{
\begin{tabular}[t]{@{}*{#1}{l@{\ }}}
#2\\ #3\\ \multicolumn{#1}{@{}l@{}}{#4}\\
\multicolumn{#1}{@{}l@{}}{#5} \\
\end{tabular}}

\newcounter{enums}

\newcounter{enumsi}

\newcommand{\enummklab}[2]
{\hspace{-.08in}#1 \hbox to 4pt{\hfil#2}} 

\newcommand{\mklab}[1]{\hfil#1}

\newcommand{\enummakelabel}[1]
{\refstepcounter{enums}  \enummklab{(\theenums)}{#1} 
  \global\let\makelabel=\mklab}

\newcommand{\ees}[1]{ 
\hspace{-.3in}
\begin{list}{\alph{enumsi}.} 
{ 
\setlength{\itemsep}{0in}   
\setlength{\topsep}{.25ex}
\setlength{\parsep}{.25ex}
\setlength{\partopsep}{.20ex}
\setlength{\labelsep}{.5em}  
\usecounter{enumsi}
\advance\leftmargin by 11pt\advance\labelwidth by 11pt%
\let\makelabel=\enummakelabel   
}
#1
\end{list}}

\newcounter{tempcnt}
\newcommand{\ex}[1]{\setcounter{tempcnt}{\value{enums}}%
\addtocounter{tempcnt}{#1}%
\arabic{tempcnt}}

\newcommand\evnup[1]{\setbox1=\hbox{#1}%
\dimen1=\ht1 \advance\dimen1 by -.5\baselineskip%
\leavevmode\lower\dimen1\box1}

\begin{document}

\maketitle
\bibliographystyle{acl}

\begin{abstract}
  In this paper, I discuss machine translation of English text into a
  relatively ``free'' word order language, specifically Turkish. I
  present algorithms that use contextual information to determine what
  the topic and the focus of each sentence should be, in order to
  generate the contextually appropriate word orders in the target language.
\end{abstract}

\section{Introduction}

Languages such as Catalan, Czech, Finnish, German, Hindi, Hungarian,
Japanese, Polish, Russian, Turkish, etc. have much freer word order
than English. For example, all six permutations of a transitive
sentence are grammatical in Turkish (although SOV is the most common).
When we translate an English text into a ``free'' word order language,
we are faced with a choice between many different word orders that are
all syntactically grammatical but are not all felicitous or
contextually appropriate.  In this paper, I discuss machine
translation (MT) of English text into Turkish and concentrate on how
to generate the appropriate word order in the target language based on
contextual information.

The most comprehensive project of this type is
presented in \cite{Stys/Zemke-95} for MT into Polish. They use the
referential form and repeated mention of items in the English text in
order to predict the salience of discourse entities and order the
Polish sentence according to this salience ranking. They also rely on
statistical data, choosing the most frequently used word orders.  I
argue for a more generative approach: a particular information
structure (IS) can be determined from the contextual information and
then can be used to generate the felicitous word order.  This paper
concentrates on how to determine the IS from contextual information
using centering, old vs. new information, and contrastiveness.
\cite{Hajicova/etal-93,Steinberger-94} present approaches that
determine the IS  by using cues such as
word order, definiteness, and complement semantic types (e.g. temporal
adjuncts vs. arguments) in the source language, English.  I believe
that we cannot rely upon cues in the source language 
in order to determine the IS of the 
translated text. Instead, I use contextual
information in the target language to determine the IS of sentences in
the target language.

In section 2, I discuss the Information Structure, and specifically
the topic and the focus in naturally occurring Turkish data. Then, in
section 3, I present algorithms for determining the topic and the
focus, and show that we can generate contextually appropriate word
orders in Turkish using these algorithms in a simple MT
implementation.

\begin{figure*}[htb]
\begin{center}
\begin{tabular}{|l|rr|}\hline
\multicolumn{3}{|c|}{\bf The Cb in SOV sentences.} \\*  \hline \hline
 Cb = Subject &            14   & (47\%) \\* \hline
 Cb = Object &          6   &  (20\%) \\* \hline
 Cb = Subj or Obj? &       6      &  (20\%) \\* \hline
 Cb = Subj or Other Obj? &  0      &  (0\%) \\* \hline
 No Cb  &               4  &  (13\%) \\* \hline
        TOTAL & 30 &    \\* \hline 
\end{tabular} \hspace{.1in}
\begin{tabular}{|l|rr|} \hline
\multicolumn{3}{|c|}{\bf The Cb in OSV sentences.} \\*  \hline \hline
 Cb = Subject &             4         & (13\%) \\* \hline
 Cb = Object &             16       &  (53\%) \\* \hline
 Cb = Subj or Obj? &       6      &  (20\%) \\* \hline
 Cb = Subj or Other Obj? &  2      &  (7\%) \\* \hline
 No Cb  &                   2  &  (7\%) \\* \hline
        TOTAL & 30 &    \\* \hline 
\end{tabular}
\end{center}
\caption{The Cb in SOV and OSV Sentences.}
\label{T-SOV-OSV}
\end{figure*}

\section{Information Structure}

In the Information Structure (IS) that I use for Turkish, a sentence
is first divided into a topic and a comment. The topic is the main element
that the sentence is about, and the comment is the information
conveyed about this topic.  
Within the comment, we find the focus, the most information-bearing
constituent in the sentence, and the ground, the
rest of the sentence.  The focus is the new or important information
in the sentence and receives prosodic prominence in speech.  

In Turkish, the pragmatic function of topic is assigned to the
sentence-initial position and the focus to the immediately preverbal
position, following \cite{Erguvanli-84}. The rest of the sentence
forms the ground.  

In \cite{Hoffman,Hoffman-tez}, I show that the information structure
components of topic and focus can be successfully used in generating
the context-appropriate answer to database queries.  Determining the
topic and focus is fairly easy in the context of a simple question,
however it is much more complicated in a text.  In the following
sections, I will describe the characteristics of topic, focus, and
ground components of the IS in naturally occurring texts analyzed in
\cite{Hoffman-tez} and allude to possible algorithms for determining
them. The algorithms will then be spelled out in section 3.

An example text from the corpus\footnote{The data was collected from
  transcribed conversations, contemporary novels, and adult speech
  from the CHILDES corpus.} is shown below.  The noncanonical OSV word
order in (\ex{1})b is contextually appropriate because the object
pronoun is a discourse-old topic that links the sentence to the
previous context, and the subject, ``your father'', is a discourse-new
focus that is being contrasted with other relatives.  {\em
  Discourse-old} entities are those that were previously mentioned in
the discourse while {\em discourse-new} entities are those that were
not \cite{Prince-92}.  \ees{
\item \gloss{6}
{Bu & defteri & de & \c{c}ok & sevdim & ben.}
{This & notebk-acc & too & much & like-pst-1S & I.}
{`As for this notebook, I like it very much.'}
\item \extragloss{5}
{Bunu & da & baban & m{\i} & verdi? (OSV)}
{This-Acc & too & father-2S & Quest & give-Past?}
{`Did your FATHER give this to you?'}
{(CHILDES 1ba.cha)}
}

Many people have suggested that ``free'' word order languages order
information from old to new information.  However, the Old-to-New
ordering principle is a generalization to which exceptions can be
found.  I believe that the order in which speakers place old vs. new
items in a sentence reflects the information structures that are
available to the speakers.  The ordering is actually the Topic
followed by the Focus. The Topic tends to be discourse-old information
and the focus discourse-new.  However, it is possible to have a
discourse-NEW topic and a discourse-OLD focus, as we will see in the
following sections, which explains the exceptions to the Old-To-New
ordering principle.



\begin{figure*}[t]
\begin{center}
\begin{tabular}{|l|rr|rr|rr|}\hline
& \multicolumn{2}{|c}{S-init} & \multicolumn{2}{|c}{IPV}
 & \multicolumn{2}{|c|}{Post-V}  \\* 
& \multicolumn{2}{|c}{{\underline S}OV,{\underline O}SV} &
\multicolumn{2}{|c}{S{\underline{O}}V,O{\underline{S}}V} &
\multicolumn{2}{|c|}{OV{\underline{S}}, SV{\underline{O}}} \\* \hline \hline
 Discourse-Old & 55 & (85\%) &    43 & (67\%) & 56 & (93\%)  \\* \hline 
  Inferrable &   8    & (13\%) &    10  & (16\%) & 4 & (7\%) \\ \hline
 D-New, Hearer-Old & 1  & (2\%) & 1 & (2\%) & 0 &  \\ \hline \hline
$\star$ D-New, Hearer-New & 0 & & 10 & (15\%) & 0 &  \\ \hline \hline
        TOTAL & 64 && 64 & & 60 &\\*  \hline 
\end{tabular} 
\end{center}
\caption{Given/New Status in Different Sentence Positions}
\end{figure*}

\subsection{Topic}

Although humans can intuitively determine what the topic of a sentence
is, the traditional definition (what the sentence
is about) is too vague to be implemented in a computational system.
I propose heuristics based on familiarity and salience 
to determine discourse-old sentence topics, and heuristics based on
grammatical relations for discourse-new topics.
Speakers can shift to a new topic 
 at the start of a new discourse segment, as in  (\ex{1})a.
Or they can continue talking about the same discourse-old topic, as in
(\ex{1})b.
\ees{
\item {[Mary]$_{T}$ went to the bookstore.}
\item {[She]$_T$ bought a new book on linguistics.}
}


A discourse-old topic often serves to link the sentence to the previous
context by evoking a familiar and salient discourse entity.  
Centering Theory \cite{GJW-95} provides a measure of saliency 
based on the
observations that salient discourse entities are often mentioned
repeatedly within a discourse segment and  are often realized
as pronouns.  \cite{Turan-95} provides a comprehensive study of null and
overt subjects in Turkish using Centering Theory, and I investigate
the interaction between word order and Centering in Turkish in 
\cite{Hoffman-center}.

In the Centering Algorithm, each utterance in a discourse is
associated with a ranked list of discourse entities called the
forward-looking centers (Cf list) that contains every discourse entity
that is realized in that utterance. The Cf list is usually
ranked according to a hierarchy of grammatical relations, e.g.
subjects are assumed to be more salient than objects. 
The backward looking center (Cb) is the most salient member of the Cf list
that links the current utterance to the previous utterance.  The Cb of
an utterance is defined as the highest ranked element of the previous
utterance's Cf list  that also occurs in 
the current utterance.  If there is a pronoun in the
sentence, it is likely to be the Cb.  As we will see, the Cb has much
in common with a sentence-topic.


The Cb analyses of the canonical SOV and the noncanonical OSV word
orders in Turkish are summarized in Figure 1 (forthcoming study in
\cite{Hoffman-center}).  As expected, the
subject is often the Cb in the SOV sentences. However, in the OSV
sentences, the object, not the subject, is most often the Cb of the
utterance.
A comparison of the 20 discourses in the first two rows\footnote{The centering analysis is
  inconclusive in some cases because the subject and the object in the sentence are
  realized with the same referential form (e.g. both as
  overt pronouns or as full NPs).} of the tables
in Figure 1 using the chi-square test shows that the association
between sentence-position and Cb is statistically significant ($\chi^2
= 10.10, \rho < 0.001$).\footnote{Alternatively, using 
 the canonical SOV sentences as the expected
  frequencies, the observed frequencies  for the
  noncanonical OSV sentences significantly diverge from the expected
  frequencies ($\chi^2 = 8.8, \rho < 0.005$).} Thus, the Cb, when it
is not dropped, is often placed in the sentence initial topic position
in Turkish regardless of whether it is the subject or the object of
the sentence.  The intuitive reason for this is that speakers want to
form a coherent discourse by immediately linking each sentence to the
previous ones by placing the Cb and discourse-old topic in the
sentence-initial position.

There are also situations where no Cb or discourse-old topic can be
found. Then, a discourse-new topic can be placed in the
sentence-initial position to start a new discourse segment.
Discourse-new topics are often subjects  or situation-setting adverbs
(e.g. yesterday, in the morning, in the garden) in Turkish.

\subsection{Focus}

The term focus has been used with many different meanings.  Focusing is
often associated with new information, but it is well-known that old
information, for example pronouns, can be focused as well. I think
part of the confusion lies in the distinction between contrastive and
presentational focus. Focusing discourse-new information is often
called presentational or informational focus as shown in
(\ex{1})a. Broad/wide focus (focus projection) is also
  possible where the rightmost element in the phrase is accented, but
  the whole phrase is in focus.
However, we can also use focusing in order to contrast one item with
another, and in this case the focus can be discourse-old or
discourse-new, e.g.  (\ex{1})b. 
\ees{
\item 
{What did Mary do this summer?} 


{She [wandered around TURKEY]$_{F}$.}

\item 
{It wasn't [ME]$_{F}$ -- It was [HER]$_F$.}
}


\cite{Vallduvi-92} defines focus as the most information-bearing
constituent, and this definition encompasses both contrastive and
presentational focusing.  I use this definition of focus as well.
However, as will see, we still need two different algorithms in order
to determine which items are in focus in the target sentence in MT. We
must check to see if they are discourse-new information as well as
checking if they are being contrasted with another item in the
discourse model.

In Turkish, items that are presentationally or contrastively focused
are placed in the immediately preverbal (IPV) position and receive the
primary accent of the phrase.\footnote{Some languages such as Greek
  and Russian treat presentational and contrastive focus differently
  in word order.} As seen in Figure 2, brand-new discourse entities
are found in the IPV position, but never in other positions in the
sentence in my Turkish corpus. The distribution of brand-new (the
starred line of the table) versus discourse-old information (the rest
of the table\footnote{{\em Inferrables} refer to entities that the
  hearer can easily accommodate based on entities already in the
  discourse model or the situation. {\em Hearer-old} entities are
  well-known to the speaker and hearer but not necessarily mentioned
  in the prior discourse \cite{Prince-92}. They both behave like
  discourse-old entities.}) is statistically significant, ($\chi^{2} =
10.847, \rho < .001$).  This supports the association of discourse-new
focus with the IPV position.

However, as can be seen in Figure 2, most of the focused subjects in 
the OSV sentences in my corpus were actually discourse-old information.
Discourse-old entities that occur in the IPV position are
contrastively focused.  In \cite{Rooth-85}'s alternative-set theory, a
contrastively focused item is interpreted by constructing a set of
alternatives from which the focused item must be distinguished.
Generalizing from his work, we can determine whether an entity {\em should}
be contrastively focused by seeing if we can construct an alternative
set from the discourse model.

\subsection{Ground}

Those items that do not play a role in IS of the sentence
as the topic or the focus form the ground of the sentence.
In Turkish, discourse-old information that is not the topic or focus 
can be 
\ees{
\item dropped,
\item postposed to the right of the verb,
\item or placed unstressed between the topic and the focus.
}
Postposing plays a backgrounding function in Turkish, and it is very
common. Often, speakers will drop only those items that are very
salient (e.g. mentioned just in the previous sentence) and postpose
the rest of the discourse-old items. However, the conditions for
dropping arguments can be very complex. \cite{Turan-95} shows
that there are semantic considerations; for instance, 
 generic objects are often dropped, but specific
objects are often realized as overt pronouns and fronted. Thus, the
conditions governing dropping and postposing are areas that
require more research.

\section{The Implementation}

In order to simplify the MT implementation, I concentrate on
translating short and simple English texts into Turkish, using an
interlingua representation where concepts in the semantic
representation map onto at most one word in the English or Turkish
lexicons.  The translation proceeds sentence by sentence (leaving
aside questions of aggregation, etc.), but contextual information is
used during the incremental generation of the target text. These
simplifications allow me to test out the algorithms for determining
the topic and the focus presented in this section.

In the implementation, first, an English sentence is parsed
with a Combinatory Categorial Grammar, CCG,
\cite{Steedman-85}.  The semantic representation is
then sent to the sentence planner for Turkish. The Sentence Planner
uses the algorithms in the following subsections to determine the
topic, focus, and ground from the given semantic representation and
the discourse model.  Then, the sentence planner sends the semantic
representation and the information structure it has determined to the
sentence realization component for Turkish.
 This component consists of a head-driven bottom up
generation algorithm that uses the semantic as well as the information
structure features given by the planner to choose an appropriate head
in the lexicon.
The grammar used for the generation of Turkish is a lexicalist
formalism called Multiset-CCG \cite{Hoffman,Hoffman-tez}, an extension of CCGs.
Multiset-CCG was developed in order to capture formal and descriptive
properties of ``free'' and restricted word order  in
simple and complex sentences (with discontinuous constituents and long
distance dependencies).  Multiset-CCG captures the context-dependent
meaning of word order in Turkish by compositionally deriving the
predicate-argument structure and the information structure of a
sentence in parallel.

The following sections describe the algorithms used by the sentence
planner to determine the IS of the Turkish sentence, given the
semantic representation of a parsed English sentence.

\subsection{The Topic Algorithm}

As each sentence is translated, we update the discourse model, and
keep track of the forward looking centers list (Cflist) of the last
processed sentence. This is simply a list of all the discourse enities
realized in that sentence ranked according to the theta-role hierarchy
found in the semantic representation. Thus, the Cf list for the
representation {\em give(Pat,Chris,book)} is the ranked list {\bf
  [Pat,Chris,book]}, where the subject is assumed to be more salient
than the objects.

Given the semantic representation for the sentence, the discourse
model of the text processed so far, and the ranked Cf lists of the
current and previous sentences in the discourse, the following
algorithm determines the topic of the sentence.  First, the algorithm
tries to choose the most salient discourse-old entity as the sentence
topic.\footnote{\cite{Stys/Zemke-95} use the saliency ranking to order
  the whole sentence in Polish. However, I believe that there is a
  distinct notion of topic and focus in Turkish.} 
If there is no discourse-old entity realized in the sentence, then a
situation-setting adverb or the subject is chosen as the discourse-new
topic.
\begin{enumerate}
\item 
Compare the current Cf list with the previous sentence's Cf list and choose
the first item that is a member of both of the ranked
lists  (the Cb).
\item 
If 1 fails: Choose the first item in the current sentence's Cf list that
is discourse-old (i.e. is already in the discourse model).
\item 
If 2 fails: If there is a situation-setting adverb in the semantic
representation (i.e. a predicate modifying the main event in
representation), choose it as the discourse-new topic. 
\item 
If 3 fails: choose the first item in the Cf list (i.e. the subject) 
as the discourse-new topic. 
\end{enumerate}

Note that the determination of the sentence topic is distinct from the
question of how to realize the salient Cb/topic (e.g. as a dropped or
overt pronoun or full NP). In the MT domain, this can be determined by
the referential form in the source text. This trick can also be used
for accommodating inferrable or hearer-old entities
that  behave as if they are  discourse-old even though
they are literally discourse-new.
If an item that is not in the discourse model is nonetheless realized
as a definite NP in the source text, 
the speaker is treating the entity as discourse-old.  This is very similar
to \cite{Stys/Zemke-95}'s  MT
system  which uses the referential form in the source text to
predict the topicality of a phrase in the target text.

\subsection{The Focus Algorithm}

Given the rest of the semantic representation for the sentence and the
discourse model of the text processed so far, the following algorithm
determines the focus of the sentence.  The first step is to determine
presentational focusing of discourse-new information.  Note that the
focus, unlike the topic, can contain more than one element; this
allows broad focus as well as narrow focusing.  If there is no
discourse-new information, the second step in the algorithm allows
contrastive focusing of discourse-old information.  In order to
construct the alternative sets, a small knowledge base is used to
determine the semantic type (agent, object, or event) of the entities
in the discourse model.
\begin{enumerate}
\item If there are any discourse-new entities (i.e. not in the
  discourse model) in the sentence, put their semantic representations
  into focus.
\item Else for each discourse entity realized in the sentence,
\begin{enumerate}
 \item Look up its semantic type in the KB and construct an
   alternative set that consists of all objects of that type in the
   discourse model,
 \item If the constructed alternative set is not empty, put
   the discourse entity's semantic representation into the focus.
\end{enumerate}
\end{enumerate}

Once the topic and focus are determined, the remainder of the semantic
representation is assigned as the ground. For now, items in the ground are
either  generated in between the topic and the focus or post-posed
behind the verb as backgrounded information.  Further research is
needed to disambiguate the use of the two possible word orders.

Further research is also needed on the exact role of verbs in the IS.
Verbs can be in the focus or the ground in Turkish; this cannot be
seen in the word order, but it is distinguished by sentential stress
for narrow focus readings.  The algorithm above works for verbs since
I place events that are realized as verbs in translated sentences into
the discourse model as discourse-old information.  However, verbs are
usually not in focus unless they are surprising or contrastive or in a
discourse-initial context.  Thus, the algorithm needs to be extended
to accommodate discourse-new verbs that are nonetheless expected in
some way into the ground component.  In addition, verbs often
participate in broad focus readings, and further research is needed to
account for the observation that broad focus readings are only
available in canonical word orders.

\subsection{Examples}

The English text in (\ex{1}) is translated using the word orders in
(\ex{2}) following the algorithms given above.  In (\ex{2}), the
numbers following T and F indicate the step in the respective
algorithm which determined the topic or focus for that sentence.
Note that the inappropriate word orders (indicated by \#) cannot be
generated by the algorithm.
\ees{
\item Pat will meet Chris today. 
\item There is a talk at four.
\item Chris is giving the talk.
\item Pat cannot come.
}

\ees{
\item \lessgloss{4}
{Bug\"{u}n & Pat & Chris'le &  bulu\c{s}acak. {\small (AdvSOV)}}
{Today & Pat & Chris-with &  meet-fut. (T:3,F:1)}  
\item \lessgloss{4}
{D\"{o}rtde &  bir & konu\c{s}ma & var. {\small (AdvSV,\#SAdvV)}}
{Four-Loc & one & talk & exist.  (T:3,F:1)}
\item \lessgloss{3}
{ Konu\c{s}may{\i} & Chris & veriyor. {\small (OSV,\#SOV)}}
{Talk-Acc & Chris & give-Prog. (T:1,F:2) }
\item \lessgloss{2}
{ Pat & gelemiyecek. ~~  (SV,\#VS)}
{Pat & come-Neg-Fut.~~(T:2,F:1 for the verb)}
}

The algorithms can also utilize long distance scrambling in Turkish,
i.e. constructions where an element of an embedded clause has been
extracted and scrambled into the matrix clause in order to play a role
in the IS of the matrix clause. For example the b sentence in the
following text is translated using long distance scrambling because
``the talk'' is the Cb of the utterance and therefore the best
sentence topic, even though it is the argument of an embedded clause.

\ees{
\item There is a talk at four.
\item Pat thinks that Chris will give the talk.
}

\ees{
\item \lessgloss{4}
{D\"{o}rtde &  bir & konu\c{s}ma & var. (AdvSV)}
{Four-Loc & one & talk & exist.  }

\item  \lessgloss{4}
{Konu\c{s}may{\i}$_{i}$ & Pat &  [Chris'in & e$_{i}$ verece\u{g}ini]}
{Talk-Acc$_{i}$ & Pat & [Chris-gen & e$_{i}$         give-ger-3s-acc]}

\lessgloss{1}
{  san{\i}yor. ~~~ (O$_{2}$S$_{1}$[S$_{2}$V$_{2}$]V$_{1}$)}
{  think-Prog. ~~~ (T:1,F:1)}
}

\section{Conclusions}

In the machine translation task from English into a ``free'' word
order language, it is crucial to choose the contextually appropriate
word order in the target language. In this paper, I discussed how
to determine the appropriate word order using contextual information
in translating into Turkish. I presented algorithms for determining
the topic and the focus of the sentence. These algorithms are
sensitive to whether the information is old or new in the discourse
model (incrementally constructed from the translated text); whether
they refer to salient entities (using Centering Theory); and
whether they can be contrasted with other entities in the discourse
model.  Once the information structure for a semantic representation
is constructed using these algorithms, the sentence with the
contextually appropriate word order is generated in the target
language using Multiset CCG, a grammar which integrates syntax and
information structure.

\bibliographystyle{/home/hoffman/tex/bibstyle}

\begin{thebibliography}{}



\bibitem[\protect\citename{Erguvanl{\i}}1984]{Erguvanli-84}
Eser~Emine Erguvanl{\i}. 1984.
\newblock {\em The Function of Word Order in Turkish Grammar}.
\newblock University of California Press. 

\bibitem[\protect\citename{Grosz/etal}1995]{GJW-95}
Barbara Grosz and Aravind K. Joshi and Scott Weinstein. 1995.
Centering: A Framework for Modelling the Local
                  Coherence of Discourse.
{\em Computational Linguistics}. 


\vspace{.5in}

\bibitem[\protect\citename{Haji\u{c}ov\'{a}/etal}1993]{Hajicova/etal-93}
Haji\u{c}ov\'{a}, Eva, Petr Sgall, and Hana Skoumalov\'{a}. 1993.
\newblock Identifying Topic and Focus by an Automatic Procedure.
\newblock {\em Proceedings of the Sixth Conference of the European
                  Chapter of the Association for Computational
                  Linguistics}. 


\bibitem[\protect\citename{Hoffman}1995]{Hoffman}
Beryl Hoffman.  1995.
\newblock Integrating Free Word Order Syntax and Information
Structure.
\newblock {\em Proceedings of the European Association for Computational
Linguistics (EACL)}. 


\bibitem[\protect\citename{Hoffman}1995b]{Hoffman-tez}
Beryl Hoffman.  1995.
\newblock {\em The Computational Analysis of the Syntax
                  and Interpretation of ``Free'' Word Order in
                  {Turkish}}.
\newblock  Ph.D. dissertation. IRCS Tech Report 95-17.
Dept. of Computer and Information Science.
University of Pennsylvania.

\bibitem[\protect\citename{Hoffman}1996]{Hoffman-center}
Beryl Hoffman. to appear 1996.
\newblock Word Order, Information Structure, and Centering in Turkish.
\newblock {\em Centering in Discourse}. 
\newblock eds. Ellen Prince, Aravind Joshi, and
Marilyn Walker. 
\newblock  Oxford University Press.


\bibitem[\protect\citename{Prince}1992]{Prince-92}
Ellen F. Prince.
\newblock The {ZPG} Letter: Subjects, Definiteness and Information
Status.
\newblock {\em Discourse description: diverse analyses of a fund
  raising text.}
\newblock eds. Thompson, S. and Mann, W. Philadelphia: John Benjamins
B.V.  pp.295--325. 1992.



\bibitem[\protect\citename{Rooth}1985]{Rooth-85}
Mats Rooth. 1985.
\newblock Association with Focus.
\newblock Ph.D. Dissertation. University of Massachusetts. Amherst.


\bibitem[\protect\citename{Steedman}1985]{Steedman-85}
Mark Steedman. 1985.
\newblock Dependencies and coordination in the grammar of {Dutch} and
                 {English},
\newblock {\em Language}, 61:523--568.


\bibitem[\protect\citename{Steinberger}1994]{Steinberger-94}
Ralf Steinberger. 1994.
\newblock Treating Free Word Order in Machine Translation.
\newblock {\em Coling}, Kyoto, Japan.

\bibitem[\protect\citename{Stys/Zemke}1995]{Stys/Zemke-95}
Malgorzata E. Stys and Stefan S. Zemke.  1995.
\newblock Incorporating Discourse Aspects in {English-Polish} {MT}:
  Towards Robust Implementation.
\newblock {\em Recent Advances in NLP}.

\bibitem[\protect\citename{Turan}1995]{Turan-95}
\"{U}mit Turan.  1995.
\newblock {\em Null vs. Overt Subjects in {Turkish} Discourse: A
            Centering Analysis}.
\newblock University of Pennsylvania, Linguistics Ph.D. dissertation. 

\bibitem[\protect\citename{Vallduv\'{\i}}1992]{Vallduvi-92}
Enric Vallduv\'{\i}. 1992.
\newblock {\em The Informational Component}.
\newblock New York: Garland. 


\end{thebibliography}

\end{document}